\documentclass[runningheads]{llncs}
\usepackage{cite}
\usepackage[inkscapeformat=png]{svg}
\usepackage{multirow}
\usepackage{booktabs}
\usepackage{datatool}
\usepackage{pgfplotstable}
\pgfplotsset{compat=1.8}

\usepackage{pgfplots} 
\pgfplotsset{compat=newest}
\usepgfplotslibrary{groupplots}
\usepgfplotslibrary{dateplot}
\usepackage{tikz}
\begin{document}
\title{Leveraging Knowledge Graph Embeddings to
Enhance Contextual Representations for Relation
Extraction}
%
%
\author{Fréjus A. A. Laleye\inst{1} \and
Loïc Rakotoson\inst{1} \and
Sylvain Massip\inst{1}} 
\authorrunning{F.A.A. Laleye et al.}
%
\institute{Opscidia, Paris, France\\
\email{\{firstname.lastname\}@opscidia.com}}
\maketitle              
\begin{abstract}

Relation extraction task is a crucial and challenging aspect of Natural Language Processing. Several methods have surfaced as of late, exhibiting notable performance in addressing the task; however, most of these approaches rely on vast amounts of data from large-scale knowledge graphs or language models pretrained on voluminous corpora.
In this paper, we hone in on the effective utilization of solely the knowledge supplied by a corpus to create a high-performing model. Our objective is to showcase that by leveraging the hierarchical structure and relational distribution of entities within a corpus without introducing external knowledge, a relation extraction model can achieve significantly enhanced performance. We therefore proposed a relation extraction approach based on the incorporation of pretrained knowledge graph embeddings at the corpus scale into the sentence-level contextual representation. We conducted a series of experiments which revealed promising and very interesting results for our proposed approach.The obtained results demonstrated an outperformance of our method compared to context-based relation extraction models.

\end{abstract}

\section{Introduction}

In recent years, there has been growing interest among researchers in transforming text into structured information for data mining purposes. The use of pre-trained language models, both general and domain-specific, has been shown to greatly improve performance on information extraction tasks \cite{peng-etal-2019-transfer, Beltagy2019SciBERTAP, Gutierrez2022ThinkingAG, pubmedbert}. However, for more specialized domains with limited data, information extraction remains a challenge. Although these models can model contextual information in text, they may not be as effective in specialized fields due to a lack of domain-specific knowledge. To address this issue, some researchers have proposed injecting external knowledge into pre-trained models, including well-supplied ontologies \cite{Liu2019KBERTEL, OstendorffBBSRG19, papaluca-etal-2022-pretrained}. However, this approach is only viable if well-annotated ontologies or knowledge graphs are available in the target domain.


The idea of combining contextual representations from large pretrained models such as BERT with semantic representation in knowledge graphs has been widely proven to improve classification tasks, as demonstrated in \cite{Ye2020DocumentAW, sta}. This same hypothesis holds true for relation extraction tasks, but only when there is a large knowledge graph available to build rich knowledge graph embeddings. However, what if such a knowledge graph is not available? In this paper, we propose an approach that leverages the global knowledge available in the training corpus to enrich contextual representations of documents and improve the performance of relation extraction based on BERT. Our approach only relies on local knowledge from the training data, making it useful for specialized domains without large specific or humanly annotated knowledge bases.


\section{Related work}




\subsection{Pretrained Languages Models}
Several contextual representation models, such as BERT \cite{DBLP:conf/naacl/DevlinCLT19}, GPT-3 \cite{winata-etal-2021-language} or Galactica \cite{https://doi.org/10.48550/arxiv.2211.09085}, have been developed to encode words in context and have shown significant improvements in various natural language processing tasks.  However, these models may not be capable of representing multiple senses of a word or capturing relational knowledge between words in context. 
For example, the word {\it geometry} can have different meanings depending on the context, such as in mathematics or chemistry \cite{https://doi.org/10.48550/arxiv.2211.09085}, and such differences may be difficult for these models to capture. 
To address this limitation, pre-trained contextual models have been developed for specific domains, such as BioBERT, which was trained on biomedical documents (PubMed abstracts and PMC full-text articles) \cite{10.1093/bioinformatics/btz682}; ClinicalBert, which was trained on clinical notes in the MIMIC-III database  \cite{Huang2019ClinicalBERTMC}, and PubMedBERT, which was trained on abstracts from PubMed and full-text articles from PMC \cite{pubmedbert}. These domain-specific models have shown promising results in classification and relation extraction tasks, indicating the importance of incorporating domain-specific knowledge in pre-training contextual models.


\subsection{External Knowledge Injection}

In order to address the aforementioned limitation, a common approach is to enhance contextual representations by incorporating external expert knowledge. Several recent works, including \cite{zhang-etal-2019-ernie, OstendorffBBSRG19, papaluca-etal-2022-pretrained},  have utilized entity and relation knowledge from external knowledge bases such as Wikidata to improve language representation. In particular, \cite{OstendorffBBSRG19} demonstrated that the addition of metadata features and knowledge graph embeddings leads to improved document classification results. Similarly, \cite{papaluca-etal-2022-pretrained} incorporated knowledge base graph embeddings pretrained on Wikidata with transformer-based language models to improve performance on the sentential Relation Extraction task. Another study \cite{Liu2019KBERTEL} experimented with injecting specialized knowledge from CN-dbpedia \cite{Xu2017CNDBpediaAN} and HowNet \cite{Dong2010HowNetAI}, which are large-scale language knowledge bases, into the language representation on twelve NLP tasks. Their findings highlight the usefulness of knowledge graphs for tasks requiring expert knowledge. However, unlike these methods, our proposed approach dynamically learns the global and hierarchical structure of the graph built on a corpus for the relation extraction task, rather than incorporating embeddings pretrained on external knowledge graphs directly into the contextual representation. Additionally, we incorporate relational information between entities learned dynamically from their hierarchical structure in the datasource, rather than entity embeddings.


It is worth noting that all state-of-the-art methods presented in the literature require access to knowledge contained in large knowledge graphs. In these approaches, the embeddings are first pretrained on a broad range of information derived from the knowledge graph, which is separate from the graph constructed from the original data, before being incorporated into contextual representations of sentences or documents. In contrast, our proposed approach is motivated by the desire to dynamically learn the global and hierarchical structure of the graph built from the data corpus for the relation extraction task. It is important to emphasize that we do not inject entity embeddings into the contextual representation, but rather incorporate relational information dynamically learned from the hierarchical structure of the data source between two entities.

\section{Proposed approach}

In this section, we describe each component that composes the improved semantic relation extraction system we propose.
\subsection{Model architecture}

\begin{figure*}
    \centering
    \includegraphics[width=0.95\textwidth]{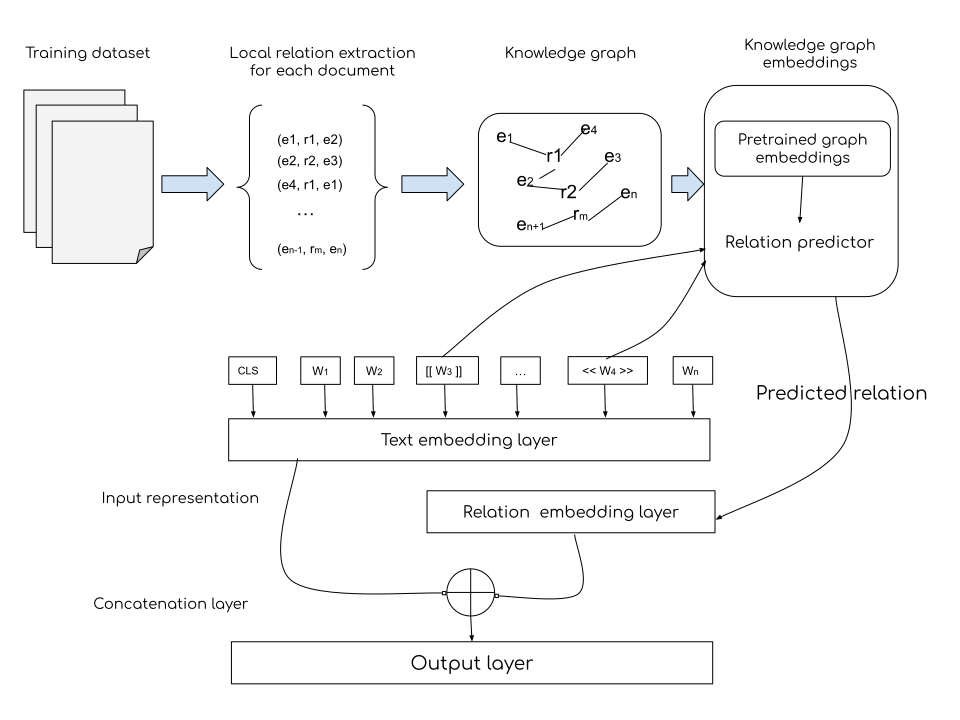}
    \caption{Architecture of our proposed approach.}
    \label{fig:arch}
\end{figure*}

We propose an approach that leverages global representations constructed from local knowledge graphs using training data independent of domain and data specialization. As illustrated in Figure \ref{fig:arch}, our model architecture consists of two main modules. The first module focuses on creating these global representations, while the second module jointly learns relational and contextual representations. Both modules are detailed in the following sections.


 In Figure \ref{fig:arch}, the collection of triples $\{ e_i, r_k, e_j \}$ designates the set of entity-relations extracted from the training documents where $e_1$ and $e_2$ represent the entities and $r$ the semantic relation  to be predicted in a given document $d$. Each triple corresponds to a document. We denote a document $d = \{w_1, w_2, w_3, ..., w_n\}$ as a sequence of tokens, where $n$ is the length of this document.
    
\subsection{Pretrained knowledge graph embeddings}

The primary goal of this module is to generate embeddings from a hierarchical knowledge graph that captures local and structural information of all entity pairs in the documents. The knowledge graph is designed to represent relational information between entities and properties between relations, allowing for the encoding and learning of relational semantic representations necessary for semantic relation prediction. To achieve this, we constructed a set of triples consisting of entities and relations found in the training data. 

Given the set of documents $D$, we constructed the graph $KG = \{h, r, t\} \subseteq E \times R \times E $
with $E$ and $R$ respectively the set of entities and the set of relations. 
Although the set $D$ only contains sentences and does not explicitly specify the knowledge graph $KG$, one can be generated by  extracting  the set of all available entities $E$ in the documents and all relations $R$ to form $KG$ whose edges are represented by $r \subseteq R$ and nodes by $\{h, t\} \subseteq E$. 

The relational representations are therefore computed by optimizing the function $f(h,t) =  hM_rt$ over the set of triples where $M_r$ is a relation-specific diagonal matrix that captures the properties of the relation between two entities. Dense vector representations (i.e., knowledge graph embeddings) are generated for each entity by training the knowledge graph, such that the distances between these vectors predict the presence of edges in the graph. Using this model, we can predict a relation $r$ given a question of the form $(e_1, ?, e_2)$, which is essentially a link prediction task in a knowledge graph. 

Our method for relation extraction does not involve masking entity types in documents, as we wish to preserve both the semantic and relational information between entity mentions. Additionally, we have not included entity types in the entity vocabulary, in line with our motivation to use only data available in the training set. It is worth noting that the knowledge graph embeddings used in previous works  \cite{joshi-etal-2020-spanbert, zhou-chen-2022-improved, lin-etal-2021-entitybert, lin-etal-2021-entitybert} are obtained from a wide range of entity properties and information drawn from external knowledge graphs, rather than directly from the graph structure provided by the training data. Our approach involves pre-training a model of knowledge graph embeddings to infer relation representations from two entities, and subsequently using this model during the training of the final semantic relation extraction system.


\subsection{BERT-based document contextuel representation}

To obtain the distributed representation of a document and encode the context for the relation extraction task, we opted for a pre-trained BERT-like model. Following the success of Transformer \cite{NIPS2017_3f5ee243}, BERT models have been experimented with in most NLP tasks and adapted in several domains. In this work, we have been interested in BERT models  such as SpanBERT \cite{joshi-etal-2020-spanbert}, a replica of BERT designed to represent and predict spans of text, SciBERT \cite{Beltagy2019SciBERTAP}, a pretrained language model for scientific text based on BERT and PubMedBERT \cite{pubmedbert}, a BERT model pre-trained with a collection of PubMed abstracts. These are all designed to learn a sequential contextual representation of the input sequence through a multi-layer, multi-head self-attention mechanism.
Specifically, the pre-trained BERT model, in our text embedding layer, is used to produce the token representation which represent the context-sensitive word embeddings. Given a document $d$, we insert the following four markers $[[$, $]]$, $<<$, and $>>$ at the beginning and end of the two entities $(e_1,e_2)$. An example from chemprot dataset is shown in Figure \ref{fig:text}

\begin{figure}
\centering
\begin{verbatim}
<< Androgen >> antagonistic effect of estramustine phosphate (EMP) metabolites
    on wild-type and mutated [[ androgen receptor ]].
\end{verbatim}
\caption{Document with special tokens.}
    \label{fig:text}
\end{figure}

\subsection{Input representation}

We built the input representation by summing the representation from BERT and relation representation obtained with the knowledge graph embeddings. The pretrained knowledge graph embedding model has the ability to assign a dense vector representing the relation that may exist between the two entities present in the text. Given a sequence of words $S = \{w_1, w_2, w_3, ... w_j\}$ as a document where $w_i$  is the $j-th$ token, including the special tokens, we built the input representation to aggregate both the contextual information and the learned relational structure.

First, in text embedding layer, we used BERT-based  model  to produce tokens embeddings $H_c$ from $S$:
\begin{equation}
    H_c = BERT(\{[CLS]\} \cup \{w_i \in S\})
\end{equation}

where $[CLS]$ is a special token that is put in the first position of each document. We obtained the contextual representation of the document by deriving its final state.

Secondly, to obtain the relation representation $H_r$, we construct the query $(e1, ?, e2)$, from the document, as input to the knowledge graph embedding model from which it infers the vector representation of $r$ :
\begin{equation}
    H_r = h*M_r*t
\end{equation}
where $h$ and $t$ are respectively the embeddings of $e1$ and $e2$ and $M_r$ the relation embedding matrix  learned during pre-training. It's worth mentioning that a limitation of our approach of using only training data to build the knowledge graph embedding model is that, with its inability to incorporate context, it can't infer an unseen relation during training. This can be a problem for the incorporation of the relation representation in the contextual vector. A simple solution that we employed is tu use a zero vector as embedding vector if no relation is found given a query. This has the advantage of always ensuring the concatenation operation with the BERT context vector.

Finally, in the concatenation layer, we construct the final representation which integrates both the contextual representation $H_c$ and the learned relation representation $H_r$:
\begin{equation}
    \nu_d = \underbrace{BERT(\{[CLS]\} \cup \{w_i \in S\})}_{H_c} + W^*\underbrace{h*M_r*t}_{H_r}
    \label{eq:nu}
\end{equation}

where $\nu_d = H_c + W^*H_r$ if the model of knowledge graph emneddings infers a relation  and $\nu_d=H_c$ otherwise. The learnable weights $W^*$ are composed of one linear layer and dropout ($p=0.1$) for regularization. Intuitively, $\nu_d$ aggregates and can fuse contextual and relational information across document tokens.

\section{Experiments}

\subsection{Dataset}

Although our approach may be domain agnostic, in this paper we focus our study of biomedical relation extraction on predicting relations between two biomedical entities given a sentence. We therefore chose to carry out experiments with the relation extraction datasets from BLURB benchmark \cite{pubmedbert}.  Table \ref{table:stats} presents the datasets and their statistics on the various training, development and testing sets.

\subsubsection{ChemProt.} ChemProt is a disease chemical biology dataset consisting of 1820 PubMed abstracts with annotated chemical-protein interactions \cite{krallinger2017overview}. The dataset contains the protein and chemicals entity types and about $10,031$ relations.

\subsubsection{DDI.} DDI dataset consists of 1025 documents from two DrugBank database and MedLine which are labeled with drug entities and drug-drug interactions \cite{HERREROZAZO2013914}. All interactions are categorized into four true and one vacuous relation.

\subsubsection{GAD} The GAD (Genetic Association Database) is an archive of published genetic association studies which consists of scientific excerpts and abstracts distantly annotated with the presence or absence gene-disease associations \cite{gad}

%
\begin{table}[!htbp]
\centering
\begin{tabular}{@{}lllll@{}}
\toprule
Dataset  & Entity type      & Train  & Dev    & Test   \\ \midrule
Chemprot & Chemical-Protein & 18,035 & 11,268 & 15,745 \\ \midrule
DDI      & Drug-Drug        & 25,296 & 2,496  & 5,716  \\ \midrule
GAD      & Gene-Disease     & 4261   & 532    & 534    \\ \bottomrule
\end{tabular}
\caption{Dataset statistics}
\label{table:stats}
\end{table}

\subsection{Baselines}

We compared our approach, which is based on the incorporation of a learned relation representation into contextualized document representations, to the following state-of-the-art context models for the semantic relation extraction task. 
\begin{enumerate}
    \item {\bf SciBERT} \cite{Beltagy2019SciBERTAP} leverages unsupervised pretraining on a large multi-domain corpus of scientific publications to improve performance on downstream scientific NLP tasks.

    \item {\bf SpanBERT}  \cite{joshi-etal-2020-spanbert} is a self-supervised pre-training method designed to better represent and predict spans of text.

    \item {\bf BioBERT} \cite{10.1093/bioinformatics/btz682} is a domain-specific language representation model pre-trained on large-scale biomedical corpora. 

    \item {\bf PubMedBERT} \cite{pubmedbert} a pretraining model using abstracts from PubMed and full-text articles from PubMedCentral that achieved state-of-the-art performance on many biomedical NLP tasks.
\end{enumerate}

These models were chosen for their application in the biomedical field and because they have proven their effectiveness in the extraction of biomedical relationship. They have all been finetuned on the 3 benchmarked datasets according to author recommendations for result reproduction. It should be mentioned that the recommendations were not enough to achieve the same performance claimed in their paper even when using their open-source code. For example, with Scibert\footnote{https://github.com/allenai/scibert}, we obtained F1-macro score of $74.2\%$ on the ChemProt dataset while the authors mentioned $83.64\%$ in their paper. In this paper, we reported the scores from our own experiments.

\subsection{Overall performance}

Once the knowledge graph is created using all entity pairs and their relations, we build the knowledge graph embedding model by using several  popular Knowledge Graph Embedding approaches such as a translation-based model TransE \cite{NIPS2013_1cecc7a7}, DistMult \cite{Yang2014EmbeddingEA} and ComplEx \cite{Trouillon2017KnowledgeGC} which are the semantic matching models. The effectiveness of each model was tested and evaluated on the three datasets and on the link prediction task. Two standard measures are used as evaluation metrics: Mean Reciprocal Rank (MRR) and Hits at $N$.
For a given entity pair, we perform a ranking of all the triples to calculate $hits@N$ and Mean Reciprocal Rank. $Hits@10$ denotes the fraction of actual triples that are returned in the top 10 predicted triples
We report MRR, and Hits at 1 and 10 in Table \ref{tab:kge} for the evaluated models. Among these KGE methods, ComplEx performed the best on the link prediction task. As a result, we only use knowledge graph embeddings from ComplEx in the following experiments.

\begin{table*}[!htpb]
\centering
\begin{tabular}{@{}cccccccccc@{}}
\multirow{2}{*}{Model} & \multicolumn{3}{c}{ChemProt} & \multicolumn{3}{c}{DDI} & \multicolumn{3}{c}{GAD} \\ \cmidrule(l){2-10} 
                       & Hits@1   & Hits@10   & MRR   & Hits@1 & Hits@10 & MRR  & Hits@1 & Hits@10 & MRR  \\ 
TransE                 & 0.44     & 0.51      & 0.49  & 0.51   & 0.56    & 0.58 & 0.31   & 0.45    & 0.38 \\ 
DistMult               & 0.52     & 0.72      & 0.59  & 0.64   & 0.77    & 0.71 & 0.37   & 0.47    & 0.45 \\ 
ComplEx                & 0.67     & 0.84      & 0.72  & 0.71   & 0.88    & 0.78 & 0.44   & 0.52    & 0.51 \\ \bottomrule
\end{tabular}
\caption{Comparing the KGE models on all the 3 datasets used in our study. We conclude an outperformance of the ComplEx model compared with the TransE and DistMult models.}
\label{tab:kge}
\end{table*}

The accuracy and F1 scores for the relation extraction models on the 3 benchmark datasets are shown in Table \ref{tab:results}. The performances of relation extraction based on contextual representations are presented at the top of the table.

\begin{table}[!htpb]
\centering

\begin{tabular}{lcccccc}
\toprule
 \multirow{2}{*}{\textbf{Model}} & \multicolumn{2}{c}{\textbf{ChemProt}} & \multicolumn{2}{c}{\textbf{DDI}} & \multicolumn{2}{c}{\textbf{GAD}} \\ \cmidrule{2-7}
  & Acc & F1& \hspace{0.5cm} Acc & F1 & \hspace{1cm} Acc &  F1  \\ \midrule
\multicolumn{7}{l}{\textbf{SOTA pretrained models}} \\ \midrule
SciBERT & 86.9 & 74.2 & \hspace{0.5cm} 87.55 & 76.10 & \hspace{1cm}76.75 & 50.60 \\
BioBERT & 88.15 & 77.13 & \hspace{0.5cm} 91.05 & 77.10 & \hspace{1cm}{\bf78.70} & {\bf55.45}  \\
PubMedBERT & {\bf 89.5} & {\bf79.63} & \hspace{0.5cm}{\bf91.65} & {\bf78.80} & \hspace{1cm}78.69 & 54.60 \\
SpanBERT & 77.20 & 65.33 & \hspace{0.5cm}80.10 & 70.95 & \hspace{1cm}76,35 & 51.80  \\
\midrule
\multicolumn{7}{l}{\textbf{Incorporating kGE-based relation representation using our proposed approach}} \\ \midrule
SciBERT+ComplEx & 90.35 & 79.45 & \hspace{0.5cm}90.65 & 80.10 & \hspace{1cm}86.45 & 59.20 \\
BioBERT+ComplEx & 92.25 & 84.33 & \hspace{0.5cm}93.65 & {\bf86.10} & \hspace{1cm}87.05 & 61.40  \\
PubMedBERT+ComplEx & {\bf93.55} & {\bf86.83} & \hspace{0.5cm}{\bf93.70} & 85.85 & \hspace{1cm}{\bf89.66} & {\bf62.82}  \\
SpanBERT+ComplEx & 88.05 & 75.33 & \hspace{0.5cm}86.55 & 77.75 & \hspace{1cm}84.75 & 58.88  \\
\bottomrule
\end{tabular}

\caption{Evaluation of binary classifiers on test data. The best results in each column are highlighted in bold font. This table provides a comparison of relation extraction performance between our proposed approach, which incorporates kGE-based relation representation, and the SOTA pretrained approach.}
\label{tab:results}
\end{table}

The results show a very significant performance of the PubMedBERT model on all 3 datasets compared to SciBERT, BioBERT and SpanBERT. We can also notice that, although SpanBERT is designed to adapt to the recognition of participating entities in a relation extraction task, it is the model that performed less well ($AVG_{F1}=62.70\%$; $AVG_{Acc}=77.88\%$) from our experiments. BioBERT and PubMedBERT gained on average, respectively $+3$ points and $+5$ points on SciBERT model. This is due to the aggregated biomedical domain knowledge in the BioBERT and PubMedBERT models compared to SciBERT. SciBERT remains a domain specific model but general for the scientific domain.

In the second part of the Table \ref{tab:results}, we have reported the performance results of our approach for incorporating relational knowledge into the contextual representation for the semantic relation extraction task. The findings demonstrate the ability of our approach to produce a textual representation that can be enriched with contextual and relational knowledge in order to increase the performance of a relation extraction model. With our approach, we significantly improved the F1 scores of the baseline models: $+9$ points for SciBERT and $+8$ points for PubMedBERT on GAD dataset, $+9$ points for BioBERT on DDI dataset and  $+10$ points for SpanBERT on ChemProt dataset. These improved performances are shown in Figures \ref{fig:acc} and \ref{fig:f1} which respectively present the average of the F1 scores and accuracies on all of the 3 datasets.

\pgfplotstableread{mean_acc.txt}{\meanacc}

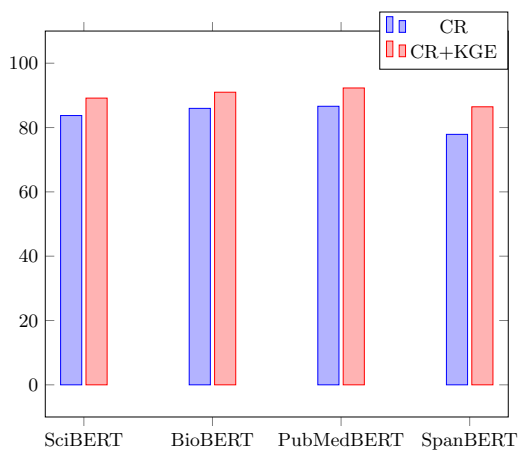
\begin{figure}[!htpb]
\centering
\begin{tikzpicture}[scale=0.8]

    \begin{axis}[
		legend entries={CR, CR+KGE},
		legend style={at={(1,0.9)},anchor=south east},
        height=8cm,
        ybar,
        ymin = 0,
		ymax = 100,
        log ticks with fixed point,
        enlargelimits=0.1,
        xtick=data,
        xticklabels from table={\meanacc}{label},
        x tick label style={anchor=north,font=\small}
    ]
        \foreach \i in {before,after}
        {\addplot table [x expr=\coordindex, y=\i] {\meanacc}; }
    \end{axis}
\end{tikzpicture}

\caption{Average of model accuracies on all 3 datasets. CR (Contextual Representation) denotes relation extraction using only contextual document embeddings and CR+KGE denotes relation extraction using our approach of incorporating knowledge graph embeddings into contextual representation.}
\label{fig:acc}
\end{figure}

\pgfplotstableread{mean_f.txt}{\meanf}

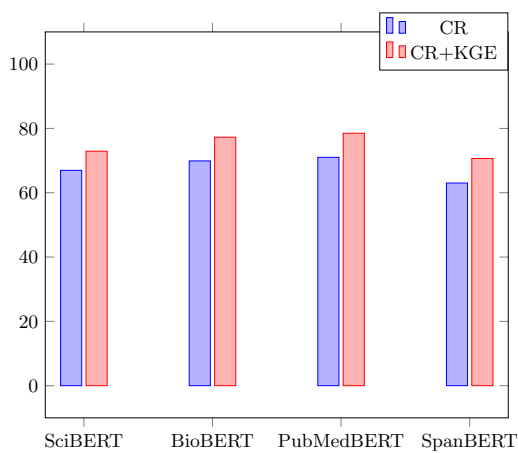
\begin{figure}[!htpb]
\centering
\begin{tikzpicture}[scale=0.8]

    \begin{axis}[
		legend entries={CR, CR+KGE},
		legend style={at={(1,0.9)},anchor=south east},
        height=8cm,
        ybar,
        ymin = 0,
		ymax = 100,
        log ticks with fixed point,
        enlargelimits=0.1,
        xtick=data,
        xticklabels from table={\meanf}{label},
        x tick label style={anchor=north,font=\small}
    ]
        \foreach \i in {before,after}
        {\addplot table [x expr=\coordindex, y=\i] {\meanf}; }
    \end{axis}
\end{tikzpicture}

\caption{Average of F1 scores on all 3 datasets. CR (Contextual Representation) denotes relation extraction using only contextual document embeddings and CR+KGE denotes relation extraction using our approach of incorporating knowledge graph embeddings into contextual representation.}
\label{fig:f1}
\end{figure}

In order to illustrate the limitations of relying solely on textual context for relation extraction, we conducted an evaluation to assess the ability of the model to generalize to documents where the relation representation provided by the Knowledge Graph Embedding (KGE) model does not align with the ground truth relations and is semantically distant from the context. To achieve this, we proposed a setting in which the test data did not include any relationships from the training set of the KGE model. Intuitively, this setting forces the relation extraction model to rely on more than just the entities, since they lack explicit links to the predicted relation. We present the evaluation results in Table \ref{tab:test}


\begin{table}[!htpb]
\centering
\begin{tabular}{@{}lcc@{}}
\toprule
Model      & Acc   & F1    \\ \midrule
SciBERT+ComplEx    & 86.7  & 69.15 \\
BioBERT+ComplEx    & 87.16 & 71.25 \\
PubMedBERT+ComplEx & 89,15 & 75.75 \\
SpanBERT+ComplEx   & 81.25 & 65.4  \\ \bottomrule
\end{tabular}
\caption{Experimental result of relation extraction on the test sets not including relations from training set.}
\label{tab:test}
\end{table}

Based on the results presented in Table \ref{tab:test}, we can confirm that our approach consistently enhances the effectiveness of relation extraction models, even when the relation representation is out of context. Our proposed method outperforms baseline approaches and validates the generalization capability of our approach to incorporate both context and relational structure to improve relation extraction models.

Furthermore, our findings demonstrate the ability of our model to capture implicit and missing relations between entities in the training set, which is made possible by leveraging the knowledge information provided by the hierarchical graph constructed in the first step. 
\section{Conclusion}


In this paper, we proposed an approach for enriching contextual representations in semantic relation extraction task by leveraging the structural and relational knowledge provided by a hierarchical knowledge graph. Our proposed pre-training task of the knowledge graph incorporates knowledge information into language representation models, resulting in a better fusion of relational and contextual information. The experimental results on the BLURB benchmark demonstrate that our approach outperforms the contextual representation models cited in the related works section by a significant margin of about $10$ points in F1 scores. Moreover, our approach is able to infer semantic relations that are not included in the pre-training dataset of the knowledge graph embeddings model, which is a significant advantage over the existing works. Our proposed approach focuses on leveraging the knowledge available in the training data and does not require pre-existing knowledge graphs, making it particularly useful for specialized domains with limited external resources. Based on our findings, pre-training the knowledge graph embeddings model enables capturing implicit knowledge related to entities, resulting in enriched contextual representations of semantic relations. These results provide a promising direction for future research in the field of semantic relation extraction.

\bibliographystyle{splncs04}
\bibliography{main}




\end{document}